\documentclass[pdflatex,sn-mathphys-num]{sn-jnl}

\usepackage{graphicx}%
\usepackage{multirow}%
\usepackage{amsmath,amssymb,amsfonts}%
\usepackage{amsthm}%
\usepackage{mathrsfs}%
\usepackage[title]{appendix}%
\usepackage{xcolor}%
\usepackage{textcomp}%
\usepackage{manyfoot}%
\usepackage{booktabs}%
\usepackage{algorithm}%
\usepackage{algorithmicx}%
\usepackage{algpseudocode}%
\usepackage{listings}%

\theoremstyle{thmstyleone}%
\theoremstyle{thmstyletwo}%

\theoremstyle{thmstylethree}%

\begin{document}

\title[Article Title]{The convergent laboratory: when AI reasoning, autonomous experiments, high performance and quantum computing reshape chemistry}


\author*[1,2]{\fnm{Eliu} \sur{Huerta}}\email{elihu@anl.gov}

\author*[3]{\fnm{Xiaoyun} \sur{Wang}}\email{xiaoyunw@nvidia.com}

\author[3]{\fnm{Geetika} \sur{Gupta}}\email{gegupta@nvidia.com}

\author[4,5]{\fnm{Edward H.} \sur{Sargent}}\email{ted.sargent@northwestern.edu}

\author[6]{\fnm{Cameron J.} \sur{Owen}}\email{cowen@lila.ai}

\author[6]{\fnm{{Victor}} \sur{{Fung}}}\email{vfung@lila.ai}

\author[7]{\fnm{Abhishek} \sur{Mitra}}\email{amitra@psiquantum.com}

\author[8,9,10]{\fnm{Austin} \sur{Cheng}}\email{austin.cheng@mail.utoronto.ca}

\author[11]{\fnm{Emma} \sur{Bouchard}}\email{ebouchar@andrew.cmu.edu}

\author[11]{\fnm{Shams} \sur{Mehdi}}\email{shamsmeh@andrew.cmu.edu}

\affil*[1]{Data Science and Learning Division, Argonne National Laboratory, Lemont, Illinois 60439, United States}
\affil[2]{Department of Computer Science, The University of Chicago, Chicago, Illinois 60637, United States}
\affil[3]{NVIDIA Corporation, 2788 San Tomas Expressway, Santa Clara, California 95051, USA}
\affil[4]{Department of Chemistry, Northwestern University, Evanston, IL, 60208 United States}
\affil[5]{Department of Electrical and Computer Engineering, Northwestern University, Evanston, IL, 60208 United States}
\affil[6]{Lila Sciences, Inc., Cambridge, Massachusetts 02142, USA}
\affil[7]{PsiQuantum, Palo Alto, California 94304, USA}
\affil[8]{Department of Computer Science, University of Toronto, Sandford Fleming Building, 10 King’s College Road, Toronto, Ontario M5S 3G4, Canada}
\affil[9]{Vector Institute for Artificial Intelligence, W1140-108 College Street, Schwartz Reisman Innovation Campus, Toronto, Ontario M5G 0C6, Canada}
\affil[10]{Department of Chemistry, University of Toronto, Lash Miller Chemical Laboratories, 80 St. George Street, Toronto, Ontario M5S 3H6, Canada}
\affil[11]{Department of Chemistry, Mellon College of Science, Carnegie Mellon University, Pittsburgh, Pennsylvania, 15213, United States}


\abstract{This Comment emerges from TPC26 (\url{https://tpc26.org}), a conference convening leaders from academia, national laboratories, and industry who are reshaping materials science discovery. The meeting explored how AI, autonomous agents, self-driving labs, higher performance and quantum computing converge to amplify their individual impact on materials science discovery. The perspectives here reflect the firsthand experiences of researchers at these frontiers and capture the essence of this global endeavor. As AI-driven reasoning, autonomous agentic frameworks, self-driving laboratories, and fault-tolerant quantum processors mature simultaneously, we offer this Comment as a reference at what we believe is a tipping point of transformative advances and productive disruption in the chemical sciences.}




\maketitle

\section*{Introduction}
\label{sec1}

Chemistry has always advanced through the interplay of theory, computation, and experiment. Today, four technological currents are converging with a force that promises to redefine this interplay entirely: large language models (LLMs) trained for scientific reasoning, self-driving laboratories (SDLs) that close the loop between hypothesis and measurement, machine-learned interatomic potentials (MLIPs) that approximate quantum accuracy at a fraction of the computational cost, fault-tolerant quantum computers (FTQCs) poised to solve electronic structure problems that remain out of reach for classical hardware, and autonomous agentic frameworks that provide the connective tissue such that all of these disparate methodological and technological advancements can be integrated into a single unified platform. Each of these capabilities is impressive in isolation. Together, they outline a new paradigm for chemical discovery in which AI agents design experiments, robots execute them, neural network potentials screen candidates at scale, and quantum processors supply the high-fidelity calibration data on which everything else depends. Critically, this paradigm is not static. Continuous retraining through reinforcement learning and online learning allows models to assimilate each new experimental outcome and simulation result, progressively sharpening predictive accuracy while compressing the time from hypothesis to validated discovery\cite{shao_deepseekmath_2024, yu_dapo_2025}. As these feedback loops accumulate domain knowledge that no single researcher could hold, they empower the exploration of compositional and structural spaces that lie beyond established chemical intuition, accelerating access to materials and molecules that conventional approaches would never prioritize. 

\section*{AI that reasons about chemistry} Recent progress in LLMs has moved well beyond text generation. NVIDIA's Nemotron~3 family consists of open models optimized for scientific reasoning and agentic workflow. The Nemotron Omni model extends reasoning capability to multi-modal scientific documents containing text, figures, and tables. As open models, Nemotron models can be fine-tuned, deployed locally, and integrated into domain-specific scientific pipelines, making them attractive for research environments that require transparency, reproducibility, and data privacy.
NeMo-RL provides the scalable reinforcement learning infrastructure underlying Nemotron post-training. It supports modern RL algorithms, including supervised fine-tuning (SFT), Direct Preference Optimization (DPO), Group Relative Policy Optimization (GRPO), and DAPO, while leveraging Megatron-Core for efficient distributed training across clusters ranging from a single GPU to hundreds of accelerators. This allows scientific reasoning models to be continuously improved using feedback generated from simulated or real scientific environments.
NeMo Gym is an open-source framework for constructing reinforcement learning environments for language agents. It separates the agent loop into modular components, including model servers, agent orchestration, task resources, tool execution, and reward verification—allowing researchers to rapidly develop domain-specific environments. This modular design enables researchers to evaluate and train models using realistic scientific workflows involving multi-step reasoning, external tools, databases, and iterative decision making. Conceptually, the ecosystem can be viewed as three complementary layers. Nemotron provides the reasoning model, NeMo Gym defines the interactive scientific environment in which the model learns, and NeMo-RL supplies the scalable optimization framework used to improve the model through reinforcement learning. Together they provide an end-to-end stack for developing scientific AI agents.~\cite{nvidia_nemotron3_2025}. On the LabBench benchmark suite, the Nemotron~3 Nano Omni model matched or exceeded GPT-4o-mini on visual scientific question answering and literature search tasks, achieving an oracle score of 89\% on literature retrieval compared with 87\% for the proprietary baseline~\cite{laurent_labbench_2024}. Crucially, these models are open-weight, enabling the scientific community to fine-tune, audit, and deploy them without dependence on commercial APIs. What makes this development consequential for chemistry is not raw benchmark performance but the training methodology. NeMo Gym pairs domain-specific RL environments, such as the Ether0 chemistry environment, with a scalable training infrastructure that supports algorithms including GRPO and DAPO on clusters ranging from a single GPU to hundreds of nodes~\cite{shao_deepseekmath_2024, yu_dapo_2025}. Reasoning budget control, implemented through custom logit processors, allows models to learn when to stop thinking, producing shorter reasoning chains with higher accuracy rather than defaulting to verbose chain-of-thought traces~\cite{muennighoff_s1_2025}. This efficiency matters: an AI agent embedded in a self-driving laboratory cannot afford to spend minutes deliberating over each decision when hundreds of experiments await scheduling. The availability of open-weight models together with open-source reinforcement learning infrastructure enables reproducible scientific AI research and facilitates community-driven improvements that would be difficult using proprietary closed APIs.

\section*{Closing the loop: self-driving laboratories} The vision of autonomous experimentation has matured from concept to operational reality. Sargent and colleagues have built a suite of seven SDLs spanning inorganic materials, organic small molecules, medicinal chemistry, polymers, formulations, organ-mimicry systems, and scale-up~\cite{macleod_sdl_2020, abolhasani_rise_2023}. In a recent demonstration targeting CO$_2$ electroreduction catalysts, the team combined robotic synthesis of metal alloy powders, parallelized membrane electrode assembly testing of 30 catalysts in two hours, and a glass-box machine learning classifier trained on both categorical composition features and thermodynamic descriptors derived from the AdsorbML pipeline~\cite{kim_sdl_joule_2025, bai_sdl_natcat_2026, lan_adsorbml_2023}. The key innovation was not automation alone but the integration of human domain expertise into the learning loop. Using the GAM Changer interface, scientists were able to inspect and directly edit the shape functions of the classifier, injecting chemical intuition about adsorption intermediates such as *CO and *CHO into the model~\cite{nori_interpretml_2019}. This ``editable glass-box'' approach cut the time to discover a record-performing propylene-selective catalyst from an estimated 60 weeks of conventional experimentation to just 5 weeks~\cite{kim_sdl_joule_2025}. SHAP analysis of the accumulated data then generated testable mechanistic hypotheses: catalysts favoring low $\Delta E_{\mathrm{*CHO}-\mathrm{*CO}}$ promoted propane via Fischer-Tropsch-like *CH$_x$ pathways, while those with high values promoted propylene through *CO dimerization and C--C coupling~\cite{lundberg_shap_2017}. At Carnegie Mellon University, the AI Science Foundry takes the SDL concept further by providing shared infrastructure at scale: 60,000 square feet of robotics-integrated laboratory space housing over 80 major instruments, unified by the Autonomous Science Acceleration Platform (ASAP). ASAP features an AI-IDE that translates natural language experimental goals into instrument commands, digital twins for pre-execution simulation and hazard checking, and mobile robots that route samples between stations. This ``cloud lab'' model addresses a bottleneck that has constrained experimental science for over a century: the pace of discovery has been set by one researcher, one bench, one experiment at a time. \section*{Machine-learned potentials: bridging scales} SDLs generate candidates; someone must evaluate them computationally before precious robot time is committed. MLIPs have emerged as the bridge between the speed of force fields and the accuracy of quantum chemistry. AIMNet2, developed by Isayev and colleagues at Carnegie Mellon, is a second-generation foundational MLIP covering common non-metal and halogen elements with hybrid-DFT-level accuracy~\cite{anstine_aimnet2_2025}. AIMNet2 predicts energies, forces, partial charges, etc. on a single GPU in seconds, scales near-linearly with system size, and enables high-throughput conformer screening that is prohibitive with density functional theory (DFT). A core design principle of AIMNet2 is embedding physics directly into the architecture, e.g., explicit charge equilibration, Coulomb interactions, and dispersion corrections rather than relying on a pure black-box. This keeps the model lightweight and fast while maintaining competitive accuracy, enabling the pretrained backbone to be fine-tuned for diverse applications such as solvation modeling, reaction network discovery, metal-catalyzed reactions, and generative design of small molecules. For example, standard MLIPs, trained on near-equilibrium geometries, fail catastrophically when bonds break and form because the atomic environments encountered along reaction coordinates are fundamentally different from those in stable conformers~\cite{anstine_aimnet2rxn_2025}. AIMNet2-Rxn addresses this gap with a task-specific model trained on bond-breaking and bond-forming operations, e.g., achieving accuracy within 0.5~kcal\,mol$^{-1}$ on the well-known Diels-Alder thermochemistry, correctly distinguishing endo versus exo selectivity and regioselectivity across 650 reactions. Another fine-tuned model, AIMNet2-Pd, extends AIMNet2's coverage to palladium-catalysed cross-coupling reactions relevant in industry, e.g., accurately characterizing geometries across the entire Suzuki-Miyaura catalytic cycle on a CPU in minutes rather than the hours required by DFT~\cite{anstine_aimnet2pd_2025}. These MLIPs do not merely approximate DFT faster; they enable qualitatively new workflows, particularly when paired with uncertainty estimation to know when the model can be trusted~\cite{mehdi2026knowing}. When an SDL needs to evaluate thousands of candidate catalysts, an MLIP can serve as the first computational filter, flagging promising compositions for robotic synthesis while routing electronically complex cases to higher-fidelity methods.

\section*{Quantum computing: the high-fidelity anchor}

Accurate prediction of molecular and materials properties within the Born--Oppenheimer approximation is based on solving the many-electron Schr\"{o}dinger equation. Within a fixed orbital basis, the exact solution is full configuration interaction (FCI), whose computational cost grows combinatorially with system size. Practical calculations therefore rely on tractable implementations of methods such as DFT, configuration interaction, coupled-cluster theory, perturbation theory, density matrix renormalization group, and quantum Monte Carlo. In practice, these implementations involve approximations, truncations, or finite-resource limitations, leading to a tradeoff between accuracy and computational cost. In this context, ``high fidelity'' refers to the ability to approach the exact solution of the chosen finite-basis electronic Hamiltonian with systematically controllable error. Fault-tolerant quantum algorithms offer a route to such FCI-quality energies, with resource requirements that scale polynomially with system size and inverse target precision~\cite{lee_thc_2021,kim_ftqc_2022}. PsiQuantum is building a utility-scale, fault-tolerant quantum computing platform based on photonic qubits, integrating semiconductor-manufactured photonic chips with optical networking, cryogenic systems, control hardware, and software. In collaboration with Boehringer Ingelheim, recent work combined block-invariant symmetry shifting (BLISS), tensor hypercontraction (THC), circuit-level optimization, and compilation for the Active Volume architecture. For active-space models of the cytochrome P450 heme group and FeMoco, this reduced estimated runtimes by factors of approximately 234 and 278, respectively, relative to the corresponding prior baselines used in the study~\cite{psi_boehringer_2025}. More recently, a complementary spectrum-amplification approach using Double-Factorized Tensor Hypercontraction (DFTHC) reduced the estimated Toffoli cost for the larger FeMoco active space by a factor of 4.3 relative to BLISS--THC~\cite{low_spectrum_amplification_2025}. This result provides a further example of how continuing advances in Hamiltonian representation and quantum algorithms can reduce the resources required for fault-tolerant quantum chemistry. These results are resource estimates for future fault-tolerant hardware rather than calculations performed on present-day quantum processors. Beyond ground-state energies, fault-tolerant algorithms have also been developed for estimating molecular observables and expectation values, and for simulating real-time chemical dynamics~\cite{steudtner_molecular_observables_2023,jornada_dynamics_2025}. These developments expand the range of chemically relevant quantities that could eventually be obtained from quantum processors. Active-space selection can also strongly affect the quality and cost of calculations for chemically realistic systems, but it is not always automated and may require expert judgment and iterative testing. PsiQuantum's RLEASE uses reinforcement learning and inexpensive orbital descriptors to enable low-cost, automated, geometry-dependent construction of compact active spaces~\cite{osaro_rlease_2026}.
High-accuracy energies, forces, and other observables could also provide training and validation data for machine-learned interatomic potentials, extending quantum-derived information to larger length and time scales. PsiQuantum's MLIPilot uses tool-calling language-model agents to automate parts of the iterative development, training, testing, and refinement of MLIPs under physically constrained validation criteria~\cite{osaro_mlipilot_2026}. Within a broader computational discovery workflow, FTQC is therefore best viewed as a specialized high-accuracy component that complements classical electronic-structure methods, machine-learned models, high-performance computing, and experiment.

\section*{Autonomous Agentic Frameworks} A critical observation underpinning the technological developments above is that there is no native interface by which these disparate modes of operation can easily interact with one another. Agentic frameworks can provide this interconnectedness, making it possible to do both autonomous in silico discovery and experimental exploration of candidates within a single framework. At Lila Sciences, agents, powered by a central reasoning model, are driving every part of the scientific method, from hypothesis generation, to autonomous in silico discovery, wet and dry lab experimentation in AI Science Factories (AISFs), and to analysis and hypothesis refinement. The framework at Lila has already been used to make discoveries in both life and physical sciences, from mRNA therapies that outperform existing commercial candidates to better catalysts for hydrogen production that will underpin the transition towards a clean energy economy. On top of being able to drive autonomous discovery of novel chemicals and materials, agentic frameworks offer the added benefit of providing robust, reproducible workflows, handling data provenance across different domains seamlessly. While each of the aforementioned technologies are impressive in their own right, full integration into an agentic framework that is able to utilize the strengths and learn the weaknesses of each in the context of the full wheel of science is where superintelligence will emerge.

\section*{Convergence: the integrated discovery stack} The power of these five capabilities is multiplicative, not additive. Consider a plausible near-term workflow for catalyst discovery. An LLM-based reasoning agent, fine-tuned with RL on chemistry environments, ingests the scientific literature and proposes candidate compositions and reaction conditions. Via an agentic framework, an MLIP screens these candidates computationally, flagging electronically straightforward cases for direct robotic testing and routing strongly correlated systems to quantum processors for high-fidelity energy and force evaluation. The resulting quantum data are automatically fed back to retrain the MLIP, progressively eliminating its blind spots. Meanwhile, an SDL synthesizes, tests, and characterizes the top candidates in parallel, streaming experimental results back to the reasoning agent, which updates its hypotheses and proposes the next iteration. The entire loop operates autonomously under the direction of the agents, with human scientists intervening not to pipette solutions but to edit model assumptions, interpret mechanistic patterns, and steer the campaign toward scientifically meaningful questions. Realizing this vision demands open infrastructure. The fact that Nemotron models are open-weight, that AIMNet2 is publicly available, that NeMo-RL and NeMo Gym are open-source, and that quantum resource estimates are published transparently is not incidental; it is essential. Chemistry cannot afford to build its future on black-box commercial services whose training data, failure modes, and update schedules are opaque. 

\section*{Outlook} We are entering an era in which the rate-limiting step in chemical discovery shifts from generating data to integrating knowledge across modalities: text, spectra, simulations, and quantum measurements. The convergent laboratory, one in which AI agents reason, robots act, neural networks approximate, and quantum processors anchor, is not a distant aspiration. Its components exist today, and the engineering challenge of connecting them is formidable but tractable, and already showing power in the form of autonomous frameworks even today. The chemistry community must engage actively in shaping this integration, ensuring that the tools remain open and accessible, the science remains reproducible, and the human scientist remains what she has always been: the one who asks the right question.

\section*{Declarations}
E.H. acknowledges support from NSF grants 
OAC-2514142 and OAC-2209892. This work was supported 
by Laboratory Directed Research and Development (LDRD) 
funding from Argonne National Laboratory, provided by 
the Director, Office of Science, of the U.S. Department
of Energy under Contract No. DE-AC02-06CH11357. E.H. led the 
writing of this article. All authors contributed to the 
writing and review of this article.

\bibliography{sn-bibliography}

\end{document}